\documentclass[letterpaper, 10 pt, conference]{ieeeconf} 
\IEEEoverridecommandlockouts
\usepackage{cite}
\usepackage{amsmath,amssymb,amsfonts}
\usepackage{algorithmic}
\usepackage{graphicx}
\usepackage{textcomp}
\usepackage{xcolor}

\usepackage{soul}
\soulregister\cite7 
\soulregister\citep7 
\soulregister\citet7 
\soulregister\ref7 
\soulregister\pageref7 

\usepackage{todonotes} 
\usepackage{multirow}
\usepackage{makecell}
\usepackage{hyperref}
\usepackage{bm}

\usepackage{float}
\usepackage{booktabs}
\usepackage{threeparttable}
\usepackage{tabularx}
\usepackage{array}

\usepackage{colortbl}
\definecolor{verylightgray}{gray}{0.95}
\newcolumntype{G}{>{\columncolor{verylightgray}}c}

\usepackage[ruled,vlined,linesnumbered]{algorithm2e}

\usepackage{cuted}
\usepackage{pifont}
\usepackage{url}
\usepackage{booktabs}

\usepackage{amsmath}
\usepackage{multirow}
\usepackage{makecell}

\usepackage{color}
\definecolor{myDarkGreen}{RGB}{0,255,0}  
\definecolor{lightblue}{RGB}{0, 176, 240}  

\begin{document}

\title{Leveraging Surgical Activity Grammar for Primary Intention Prediction in Laparoscopy Procedures
}

\author{Jie Zhang$^{1}$, Song Zhou$^{1}$, Yiwei Wang$^{1,2\dagger}$, Chidan Wan$^{3}$, Huan Zhao$^{1}$, Xiong Cai$^{3}$ and Han Ding$^{1}$
\thanks{*This work was supported in part by the National Science Foundation of China under Grant 62203180, in part by the Hubei Science and Technology Major Program under Grant 2023BCA002 and Grant 2023BAA016, in part by the Wuhan Science and Technology Major Special Project under Grant 2021022002023426, and in part by the Taihu Lake lnnovation Fund for Future Technology, HUST: 2023-A-2.}
\thanks{$^{1}$Jie Zhang, Song Zhou, Yiwei Wang, Huan Zhao, and Han Ding are with State Key Laboratory of Intelligent Manufacturing Equipment and Technology, Huazhong University of Science and Technology, Luoyu Road 1037, Wuhan 430074, China.
        {\tt\small jiezh@hust.edu.cn, zhsong@hust.edu.cn, wang\_yiwei@hust.edu.cn, huanzhao@hust.edu.cn, dinghan@hust.edu.cn}}%
\thanks{$^{2}$Yiwei Wang is also with Institute of Medical Equipment Science and Engineering, Huazhong University of Science and Technology, Luoyu Road 1037, Wuhan 430074, China.}%
\thanks{$^{3}$Chidan Wan and Xiong Cai are with Department of Hepatobiliary Surgery, Union Hospital, Tongji Medical College, Huazhong University of Science and Technology, 1277 Jiefang Ave., Wuhan 430022, China.
        {\tt\small wanchidan@hust.edu.cn, caixiong@hust.edu.cn}}%
\thanks{$^{\dagger}$ Corresponding author}
}

\maketitle

\begin{abstract}
Surgical procedures are inherently complex and dynamic, with intricate dependencies and various execution paths. Accurate identification of the intentions behind critical actions, referred to as \textit{Primary Intentions (PIs)}, is crucial to understanding and planning the procedure.
This paper presents a novel framework that advances PI recognition in instructional videos by combining top-down grammatical structure with bottom-up visual cues. 
The grammatical structure is based on a rich corpus of surgical procedures, offering a hierarchical perspective on surgical activities. A grammar parser, utilizing the surgical activity grammar, processes visual data obtained from laparoscopic images through surgical action detectors, ensuring a more precise interpretation of the visual information.
Experimental results on the benchmark dataset demonstrate that our method outperforms existing surgical activity detectors that rely solely on visual features. Our research provides a promising foundation for developing advanced robotic surgical systems with enhanced planning and automation capabilities.
\end{abstract}


\section{Introduction}
\label{sec:intro}

Surgical procedures are intricate and dynamic processes composed of interconnected steps that do not follow a strictly linear path but instead vary according to specific circumstances, patient needs, and the surgeon's judgment~\cite{nagasinghe2024not}. Despite this variability, each procedure typically follows a primary sequence of actions~\cite{zhang2022automatic}. This sequence is guided by \textbf{primary intentions (PIs)}--the fundamental goals that drive the surgery progression~\cite{ashutosh2024video}. For example, in Cholecystotomy, PIs can include dissecting the fatty tissue, cutting the cystic duct, and removing the gallbladder.
By focusing on PIs, complex procedures can be broken down into smaller, more manageable components. This breakdown enables robots to effectively learn surgical techniques from instructional videos~\cite{chang2020procedure}. Accurate identification and application of primary intentions is crucial for advancing surgical robotics. This enables the development of more sophisticated automated systems that can better understand and replicate surgical techniques, leading to improved precision and efficiency in robotic-assisted surgeries~\cite{fu2024multi,zhang2023laparoscopic}.

\begin{figure}[t]
\centering 
\includegraphics[width=1\linewidth]{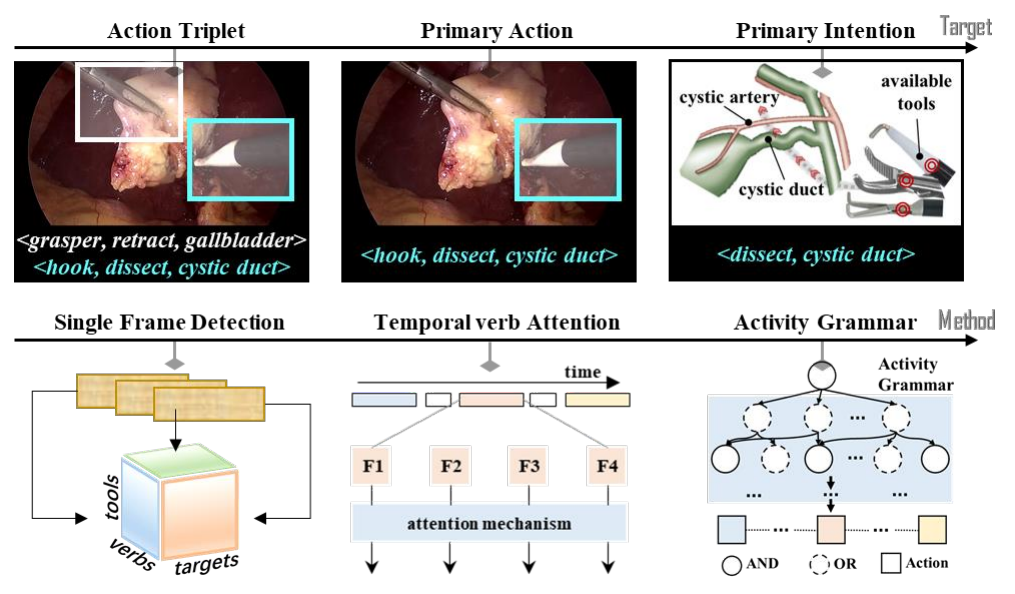} \vspace{-0.1in}
\caption{Top: the research target from broader action triplets \textless tool, verb, target\textgreater ~(left) to more specific primary intentions (center), and to the proposed primary intentions (PIs) (right). Bottom: recognition methods from using each single frame (left) to integrating continuous verb information (center), and to the proposed top-down method based on the surgical activity grammar (right).} \vspace{-0.25in}
\label{Fig. triplet_detection_modification}  
\end{figure}

For instance, laparoscopic cholecystectomy (LC) is considered the gold standard in the surgical management of gallstone disease~\cite{twinanda2016endonet}. In the LC procedure, surgical actions are typically defined as triplets in the form of \textless tool, verb, target\textgreater ~\cite{nwoye2020recognition}. As shown in the top of Fig.~\ref{Fig. triplet_detection_modification}, these triplets can be classified into \textit{primary actions}, which are necessary steps to advance the procedure (e.g., \textless hook, dissect, cystic duct\textgreater ), and \textit{auxiliary actions} that assist the primary actions (e.g., \textless grasper, retract, gallbladder\textgreater ). In real surgical scenarios, surgeons may use different tools to achieve the same operation on a target tissue. For example, both the \textless grasper, dissect, cystic duct\textgreater ~and \textless hook, dissect, cystic duct\textgreater ~triplets can be used to perform the operation \textless dissect, gallbladder\textgreater ~operation. Therefore, PI in the LC procedure can be represented as a tuple of the desired verb and the target tissue, formatted as \textless verb, target\textgreater. These PIs, which are a specialized subset of surgical actions, are essential to guide the step-by-step progression of the LC procedure.
Earlier research on surgical activity analysis was dominated by surgical phase identification. Recent studies have shown a trend from coarse-grained phase recognition to fine-grained action recognition \cite{nwoye2023cholectriplet2022}. Nwoye et al. \cite{nwoye2020recognition, nwoye2022rendezvous} introduced multi-task models to extract triplets from single frames. Building on this, Sharma et al. \cite{sharma2023rendezvous} incorporated temporal information by blending current and past verb features at the frame level, as illustrated at the bottom of Fig.~\ref{Fig. triplet_detection_modification}.
Existing approaches mainly rely on deep neural networks (DNNs) to extract visual features for action classification. However, relying solely on visual features presents several challenges:
1) \textit{Feature extraction}: DNNs may struggle to extract valid features from complex surgical videos due to factors, such as occlusions, photometric distortions, and ambiguities~\cite{nwoye2022rendezvous}.
2) \textit{Class imbalance}: The imbalanced distribution of action instances may lead models to prioritize common actions, while overlooking rare yet critical ones~\cite{nwoye2022rendezvous}. 
Consequently, bottom-up inference based solely on visual features is inadequate for PI recognition. 
Since PIs are proposed to support the understanding and planning of surgical procedures, incorporating the underlying relationships between surgical activities, such as hierarchical structure and temporal dependencies, can provide a valuable top-down perspective. This top-down approach can improve PI recognition by leveraging the contextual information provided by the overall surgical workflow.

Surgical procedures can be deconstructed into hierarchical structures comprising phases, steps, and actions, each following specific temporal patterns~\cite{hutchinson2023evaluating}. The hierarchical structure of these surgical activities is analogous to the grammar of language, where grammatical rules define the structure of sentences~\cite{kuehne2014language}. 
Previous research has demonstrated the effectiveness of grammar-based models in capturing the structure of human activities~\cite{qi2018generalized}.
Motivated by this, we propose a novel framework that combines DNN-based models with a surgical activity grammar for PI recognition. Our approach comprises three stages: 
1) A top-down model predicts the likelihood of different PIs across video frames.
2) A grammar extraction algorithm learns the underlying structure of surgical procedures from a corpus of surgical activities.
3) A grammar parser combines predicted probabilities with the learned grammar to determine the most probable PI sequence. 
By integrating top-down grammar knowledge with bottom-up visual information from images, our approach offers a more comprehensive understanding of PIs within the surgical procedure compared to methods relying solely on visual features. 
To validate our method, we performed experiments using the PI dataset, which is constructed upon the publicly available CholecT50 dataset \cite{nwoye2022rendezvous}. The results demonstrate that our grammar-augmented framework significantly outperforms state-of-the-art (SOTA) methods in PI recognition.

Our key contributions are threefold: 

1) We introduced PIs to facilitate the breakdown of complex procedures into manageable activities for robot execution, thus advancing surgical procedure automation.

2) We developed a comprehensive framework that integrates DNNs with a surgical activity grammar. This approach effectively leverages top-down grammatical insights and bottom-up visual data, bringing significantly improvements in PI recognition.

3) We built the CholecPI dataset upon the benchmark dataset and demonstrate through experiments that our proposed framework significantly outperforms SOTA methods in recognizing the PI sequence of the surgical procedure.


\section{Related works} \label{Related work} 

\subsection{Surgical Action recognition}


The components of the LC procedure can be categorized from coarse to fine granularity: phase, step, task, and action. Recognizing fine-grained surgical actions is crucial for skill assessment and automation~\cite{hutchinson2023evaluating}. 
An surgical action is commonly represented as a triplet in the format \textless instrument, verb, target\textgreater ~\cite{nwoye2022rendezvous}.
While previous research has largely focused on recognizing all intra-operative actions~\cite{nwoye2023cholectriplet2022}, this paper prioritizes the primary actions that advance the procedure. We define the recognition target as PIs, the underlying objectives of primary actions for surgical procedure planning.

Existing methods like Tripnet~\cite{nwoye2020recognition} and RDV~\cite{nwoye2022rendezvous} utilized an instrument-centric multi-task model to capture the triplet from the single frame. 
As an extension of RDV, RiT~\cite{sharma2023rendezvous} encompassed the temporal domain to blend the current and past verb features at the frame level.
While these methods can use DNNs to detect surgical action triplets from visual features, they struggle to overcome the complexities inherent in laparoscopic video data~\cite{cheng2023deep}. 
We hypothesize that surgical activity has a hierarchical structure similar to grammar in natural language~\cite{kuehne2014language}. Therefore, we propose a framework that combines DNN-based methods with a stochastic grammar model of surgical activity to improve PI recognition.

\subsection{Activity grammar}

Grammars are powerful tools for representing compositional structures. They have been extensively studied in natural language processing (NLP)~\cite{solan2005unsupervised}. As researchers explore their representational capabilities, grammars have been applied in various domains, such as human activity recognition~\cite{piergiovanni2020differentiable,qi2018generalized}, pose estimation~\cite{fang2018learning}, and vision-language induction~\cite{wan2022unsupervised}. 

These grammatical frameworks are particularly well-suited for modeling the hierarchical structure of human activities. Research has shown that grammars can effectively capture the hierarchical nature of human activities~\cite{kuehne2014language,richard2017weakly,vo2014stochastic}. For example, context-free grammars have been used to represent temporal action transitions~\cite{kuehne2017weakly,richard2018neuralnetwork}. Moreover, stochastic AND-OR grammars can hierarchically decompose complex activities into sub-events~\cite{vo2014stochastic}. 
Grammars have also been learned from action datasets~\cite{qi2017predicting,qi2018generalized} and applied in temporal action segmentation~\cite{richard2017weakly,gong2024activity}.

Building upon this grammar-based approach for human activity modeling, we represent the structure of surgical activities as an And-Or Graph (AOG). An AOG is a stochastic grammar that hierarchically decomposes the surgical process from broad phases to specific actions~\cite{kuehne2014language}.

\section{Methodology} \label{Methodology}  

\subsection{Preliminaries of activity grammar}

\begin{figure*}[t] 
\centering 
\includegraphics[width=0.9\linewidth]{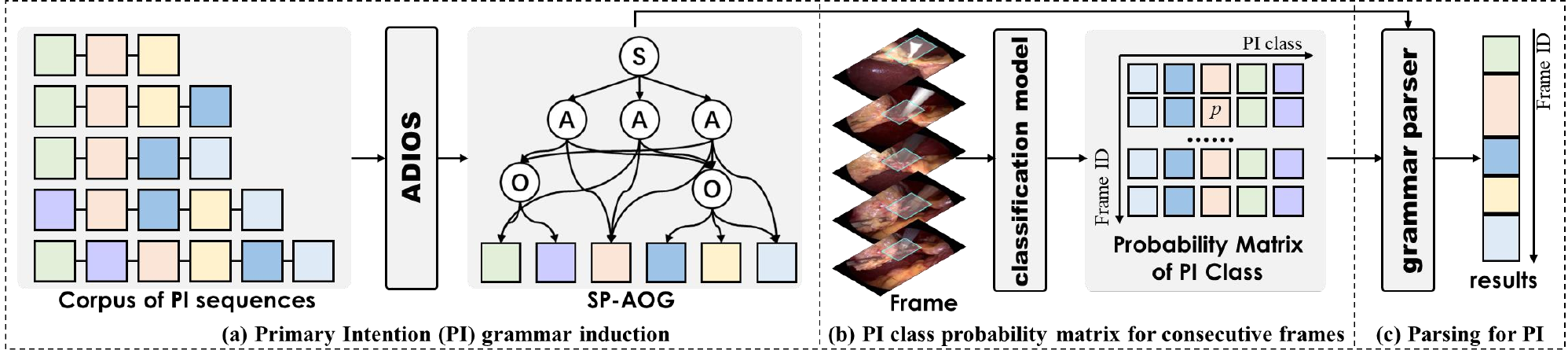} \vspace{-0.1in}
\caption{The proposed framework for PI recognition. 
(a) The surgical activity grammar, SP-AOG, is developed by statistical learning from a corpus of surgical procedure recordings, highlighting the hierarchical relationships and dependencies among PIs.
(b) SP-AOG is then used to parse a probability matrix, generated by a classification model, that indicates the likelihood of each PI category. (c) The parsing process identifies the optimal sequence of PIs that aligns with the grammar and predicted probabilities.} \vspace{-0.25in}
\label{Fig. framework} 
\end{figure*}



%


Inspired by the hierarchical structure of language, we propose the Surgical Process And-Or Graph (SP-AOG), a stochastic grammar-based model that represents the hierarchical relationships between surgical activities~\cite{kuehne2014language}. 
As shown in Fig.~\ref{Fig. framework}, SP-AOG is a probabilistic context-free grammar (PCFG) represented by a 6-tuple $G = (S, V_N, V_T, {R}, P, \Sigma)$, which models the hierarchical decomposition of surgical phases into actions through terminal and non-terminal nodes. Terminal nodes represent observed action atoms in an image, while non-terminal nodes encode grammar rules.

\begin{itemize}
    \item $S$ represents a specific surgical phase. 
    \item The non-terminal nodes $\mathrm{V_N=V_{and}\cup V_{or}}$. The \textit{And-node} $\mathrm{V_{and}}$ represents a decomposition of a large activity entity into its sub-activity constituents. 
    The \textit{Or-node} $\mathrm{V_{or}}$ has multiple mutually exclusive configurations represented by its child nodes.  
    \item $V_T$ is the set of terminal nodes that represent the actions.
     \item $R$ is a set of production rules used for top-down sampling from parent to child nodes.
    \item $P$ is the probability associated with the production rules.
    \item $\Sigma$ is the set of all valid action sequences that can be generated by the grammar.
\end{itemize}


A temporal parse tree, $pt$, is a sub-graph of the SP-AOG that captures the temporal structure within the surgical phase. The terminal nodes of $pt$ represent a set of PIs that can be performed in a valid order.



\subsection{Involving grammar into classification model}
The goal of PI recognition is to find the PI sequence that best represents the visual content and activity grammar with maximum posterior probability, 
\begin{equation}
\mathbf{A} ^*=\underset{\mathbf{A} }{\arg\max}\,p(\mathbf{A} |\mathbf{F} ,G)
\label{eq. origional_goal}
\end{equation} 
where $\mathbf{F} = \{ f_1, f_2, \cdots, f_T \}$ represents the frame sequence with length $T$, $\mathbf{A} = \{ a_1, a_2, \cdots, a_T \}$ denotes the the corresponding PI labels of $\mathbf{F}$. $G$ refers to the activity grammar for the entire surgical phase.

Due to the complex relationships between $\mathbf{A}$, $\mathbf{F}$, and $G$, optimizing the original probability model to obtain $\mathbf{A}^*$ is computationally intractable. To simplify the optimization problem, we introduce intermediate variables $\mathcal{P}$ and $\mathcal{G}$ while preserving following probabilistic semantics, 
\begin{equation}
p(\mathbf{A} |\mathbf{F} ,G) \propto  p(\mathbf{A} |\mathbf{\mathcal{P}} ,\mathcal{G})p(\mathcal{P}|\mathbf{F})p(\mathcal{G}|G)
\label{eq. transformed_goal}
\end{equation} 

According to Eq.~\ref{eq. transformed_goal}, the original problem can be approximated by solving a tractable series of sub-problems,
\begin{equation}
\begin{aligned}
\mathcal{P} ^*&=\underset{\mathcal{P}}{\arg\max}\,p(\mathcal{P}|\mathbf{A})\\
\mathcal{G} ^*&=\underset{\mathcal{G} }{\arg\max}\,p(\mathcal{G}|\mathbf{A})\\
\mathbf{A} ^*&=\underset{\mathbf{A} }{\arg\max}\,p(\mathbf{A}|\mathcal{P} ^*,\mathcal{G} ^*)\\
\end{aligned}
\label{eq. subproblems}
\end{equation} 


To solve sub-problems in Eq.~\ref{eq. subproblems}, we use the Generalized Earley Parser (GEP) method, originally developed for human activity parsing~\cite{qi2020generalized}. GEP utilizes the human activity grammar \( G_h \) to determine the most probable human action sentence, denoted as $l^*$, by analyzing a probability matrix of action sequences, $\mathbf{P}_h$,
\begin{equation}
    l^*=\underset{l}{\arg\max}\,p(l|\mathbf{P}_h,G_h)
    \label{eq. GEP}
\end{equation} 

Referring to Eq.~\ref{eq. GEP}, the solution process of sub-problems in Eq.~\ref{eq. subproblems} is specified as: 
1) A DNNs-based classification model is applied to frames $\mathbf{F}$ to obtain a category probability matrix $\mathcal{P} ^*$; 
2) The Automatic Distillation of Structure (ADIOS) algorithm~\cite{qi2017predicting} is used to learn the grammar $\mathcal{G} ^*$ from raw sequential PI corpus $\mathcal{C}$;
3) GEP processes $\mathcal{P} ^*$ as input and outputs $\mathbf{A} ^*$, which aligns with the Chomsky normal form of $\mathcal{G} ^*$.
This process can be expressed as,
\begin{equation}
\begin{aligned}
\mathcal{P} ^*&=f_{\mathrm{DNN}} (\mathbf{F})\\
\mathcal{G} ^*&=f_{\mathrm{ADIOS}} (\mathcal{C})\\
\mathbf{A} ^*&=f_{\mathrm{GEP}} (\mathcal{P} ^*|\mathcal{G} ^*)
\end{aligned}
\label{eq. solving subproblems}
\end{equation}

Through the aforementioned process, the challenging Eq.~\ref{eq. origional_goal} is transformed into the more manageable Eq.~\ref{eq. solving subproblems}.

\subsection{Primary Intention Recognition}





According to Eq.~\ref{eq. solving subproblems}, we follow three steps to optimize PI recognition by integrating surgical activity grammar: First, generate the multi-class probability matrix $\mathcal{P}^*$ for frame sequences. Next, learn the activity grammar $\mathcal{G}^*$ from the PI label corpus. Finally, generate PI sequences $\mathbf{A}^*$ based on $\mathcal{P}^*$ and $\mathcal{G}^*$.


\subsubsection{Probability matrix acquisition}
Triplet recognition is a multi-class classification problem, with the classifier $f_{\mathrm{RDV}}(\cdot)$ consisting of an encoder and classification heads. For an image $f_i$, the feature embedding $\mathbf{h}_i$ is generated after the encoder processing. Employing \textbf{SoftMax} as the activation function in the final layer is a common approach to transform $\mathbf{h}_i$ into probabilities $\mathbf{p}_i$ for multi-class classification of $f_i$, 
\begin{equation}
 \mathbf{p}_i = \begin{pmatrix}P(a_i = 1|\mathbf{h}_i,\mathbf{W})\\P(a_i = 2|\mathbf{h}_i,\mathbf{W})\\\cdots\\P(a_i = K|\mathbf{h}_i,\mathbf{W})\end{pmatrix} = \frac{1}{\sum_{k = 1}^{K}e^{-\mathbf{W}_k\mathbf{h}_i}}\begin{pmatrix}e^{-\mathbf{W}_1\mathbf{h}_i}\\e^{-\mathbf{W}_2\mathbf{h}_i}\\\cdots\\e^{-\mathbf{W}_K\mathbf{h}_i}\end{pmatrix}
\label{eq. label probability}
\end{equation}
where $K$ represents the number of PI categories, $\mathbf{W}=(\mathbf{W}_1,\mathbf{W}_2,\ldots,\mathbf{W}_K)^\top $ represents the weight vector, $a_i$ represents the predicted label of $f_i$. 
Accordingly, the multi-class probability matrix $\mathcal{P} ^*$ for the frame sequence $\mathbf{F}$ is,
\begin{equation}
    \mathcal{P} ^* = (\mathbf{p}_1, \mathbf{p}_2,\cdots ,\mathbf{p}_{T})^{\top }
    \label{eq. probability matrix}
\end{equation}
where \( T \) denotes the length of the frame sequence.

\subsubsection{SP-AOG learning}
Learning the PCFG from a corpus involves capturing the statistical patterns of syntactic structures in the text data~\cite{solan2005unsupervised}. 
SP-AOG is a PCFG proposed to represent surgical activities. To develop its grammar, the corpus $\mathcal{C}$ should be structured as the set of the PI sequence within each surgical procedure, which is, 
\begin{equation}
\begin{aligned}
c_n & = \{  \text{\small{SIL}}, {a_1} , {a_2}, \cdots ,{a}_{T},\text{\small{SIL}}\} \\
\mathcal{C} & = \{  c_1, c_2, \cdots ,c_N \} 
\end{aligned} \label{eq. corpus}
\end{equation}
where $c_n$ is a sentence that includes the PI sequence $\{{a_1}, {a_2}, \cdots, {a_T}\}$. The symbol $\text{SIL}$, denoting the silence of the action, is added at the beginning and end of the sequence. $\mathcal{C}$ consists of $N$ sentences obtained from $N$ cases of surgery.

For a sentence $c_n$, let $\mathcal{T}(c_n)$ denote all possible parse trees. The probability of observing $c_n$ under the PCFG $G$ is calculated by summing the probabilities of all parse trees in $\mathcal{T}(c_n)$,
\begin{equation}
    P(c_n | G) = \sum_{pt \in \mathcal{T}(c_n)} P(c_n, pt | G) 
\end{equation}
where $P(c_n, pt | G) $ is the joint probability of sentence $c_n $ and parse tree $pt$ under $G$.
Accordingly, $\mathcal{G} ^*$ can be learned by maximizing the likelihood of the corpus,
\begin{equation}
     \mathcal{G} ^* = \underset{G}{\arg\max}\, \sum_{n=1}^{N} \log \left( \sum_{pt \in \mathcal{T}(c_n)} P(c_n, pt | G) \right ) 
\end{equation}

\subsubsection{Primary Intention parse via GEP}
Parsing PI sequences using GEP involves analyzing the input sequence within the context of the PCFG. 
GEP produces a set of parse trees, denoted as $\mathcal{T}$. According to the learned $\mathcal{G}^*$, these trees represent all possible syntactic interpretations of the input probability matrix $\mathcal{P}^*$ for the PI sequence.
Our objective is to find the most likely PI sequence $\mathbf{A} ^*$ generated by the most probable parse tree. Therefore, we first calculate the probability of the parse tree,
\begin{equation}
    P(pt | \mathcal{G} ^*) = \prod_{r \in R(pt)} P(r | \mathcal{G} ^*)
\end{equation}
where $ R(pt) $ denotes the set of production rules used in parse tree $ pt $. $P(r | \mathcal{G} ^*) $ is the probability of the production rule $ r $ given $\mathcal{G} ^* $.
Then $\mathbf{A} ^*$ can be parsed by maximizing the likelihood of producing the PI sequence,
\begin{equation} 
\mathbf{A} ^* = \underset{\mathbf{A}}{\arg\max}\, \sum_{pt \in \mathcal{T}} P(pt | \mathcal{G} ^*) 
\end{equation}
where $\mathcal{T}$ represents the set of parse trees for the input probability matrix $\mathcal{P}^*$, which are consistent with $\mathcal{G}^*$.


To sum up, the framework starts with a DNN-based model that uses visual features to create a probability matrix for the PI sequence, which belongs to the bottom-up technique. Next, the grammar parser uses this matrix for top-down parsing to interpret the activity grammar in SP-AOG. Bottom-up and top-down processes are integrated to improve PI recognition performance.

\section{Experiments and results} \label{Experiment}
\subsection{Experimental setup}

\subsubsection{Dataset}
We introduce the \textbf{CholecPI} dataset, which is developed from the CholecT50 dataset~\cite{nwoye2022rendezvous}. The CholecT50 dataset includes 100 triplet categories distributed over 7 labeled surgical phases. Focusing on the Calot triangle dissection phase, we identified six types of PIs in the \textless verb, target\textgreater ~format after consulting expert surgeons. 
The CholecPI dataset consists of 25,380 frames, with detailed information for each PI category listed in Table~\ref{tab:dataset}.
To evaluate our model, we employ a 5-fold cross-validation approach, following the methodology in~\cite{nwoye2022rendezvous}. The 50 videos are divided into five equal-sized folds, each containing 10 videos.


\begin{table} [h] 
    \centering
    \caption{Description of the CholecPI Dataset.} \vspace{-0.1in}
    \label{tab:dataset}
    \begin{threeparttable}
    \begin{tabular}{p{0.8cm}p{3cm}p{1.2cm}p{1.2cm}}
    \toprule
    \textbf{ID}& \textbf{PI Category}  & \textbf{Number} & \textbf{Percentage}\\
    \midrule
     \textbf{PI}0 & \textit{\textless dissect, cystic\_duct\textgreater}  & 7677 & 30.25\%\\
      \textbf{PI}1 & \textit{\textless aspirate, fluid\textgreater}   & 525 & 2.07\% \\
      \textbf{PI}2 & \textit{\textless dissect, gallbladder\textgreater}   & 10459 & 41.21\% \\
      \textbf{PI}3 & \textit{\textless dissect, cystic\_artery\textgreater}   & 3059 & 12.05\% \\
     \textbf{PI}4 &  \textit{\textless dissect, cystic\_plate\textgreater}   & 2904 & 11.44\% \\
      \textbf{PI}5 & \textit{\textless dissect, fat\textgreater}$^\dagger$  & 756 & 2.98\% \\
    \bottomrule
    \end{tabular}
    \begin{tablenotes}
        \footnotesize
        \item[$\dagger$] original \textit{\textless dissect, peritoneum\textgreater}, \textit{\textless dissect, omentum\textgreater}, and \textit{\textless cut, peritoneum\textgreater} are all categorized as \textit{\textless dissect, fat\textgreater}.  
      \end{tablenotes}
    \end{threeparttable} 
\end{table}

\subsubsection{Metric}

Given the class-imbalance challenges of the CholecPI dataset, we use following evaluation metrics to assess the proposed method: \textbf{micro accuracy (Precision/Recall)}, \textbf{macro precision}, \textbf{macro recall}, \textbf{macro F1 score}, and \textbf{weighted F1 score}. Micro accuracy reflects the overall percentage of correctly classified labels across all instances. Macro precision and macro recall calculate the average precision and recall across all classes, treating each class equally. The macro F1 score is the harmonic mean of macro precision and macro recall, providing a balanced measure of performance across all classes. 
The weighted F1 score assigns greater importance to less common classes, ensuring that model performance is not dominated by more frequent classes.

\subsubsection{Baseline method}
We compare our proposed method with SOTA surgical action detectors, which are solely based on visual features.

\textbf{1) Tripnet}~\cite{nwoye2020recognition}: A weakly supervised detection model that utilizes a 3D interacting space to recognize action triplets from each single frame.

\textbf{2) Rendezvous (RDV)}~\cite{nwoye2022rendezvous}: Build upon Tripnet, RDV incorporates a transformer-inspired semantic attention module to aggregate visual features in the spatial domain.

\textbf{3) Rendezvous in Time (RiT)}~\cite{sharma2023rendezvous}: RiT extends RDV by incorporating a temporal attention module that integrates historical verb features at the frame level.

The \textless verb, target\textgreater ~classification heads are added to these baseline models to support our PI recognition task. 
To mitigate the imbalanced classification problem, we apply a weighted loss by scaling the loss for the two least frequent categories, PI0 and PI5~\cite{phan2020resolving}. The baseline models trained using this weighted loss method are denoted as \textbf{Tripnet{\tiny W}}, \textbf{RiT{\tiny W}}, and \textbf{RDV{\tiny W}}. 

\subsubsection{Implementation details}
The category probability matrices are extracted from the softmax layer. 
We randomly select 10 and 20 surgeries from the training set to learn the surgical PCFG.
CholecPI provides frame-wise annotations for verbs and target tissues. To generate segment-level PI annotations, we consider each combination of a verb label and its corresponding target as a unit. A new segment annotation is assigned when either the verb or target changes. 
Following the format in Eq.~\ref{eq. corpus}, the segment-level PI annotations can yield a corpus, which records the PI sequences. 

\begin{figure}[t] 
\centering 
\includegraphics[width=1\linewidth]{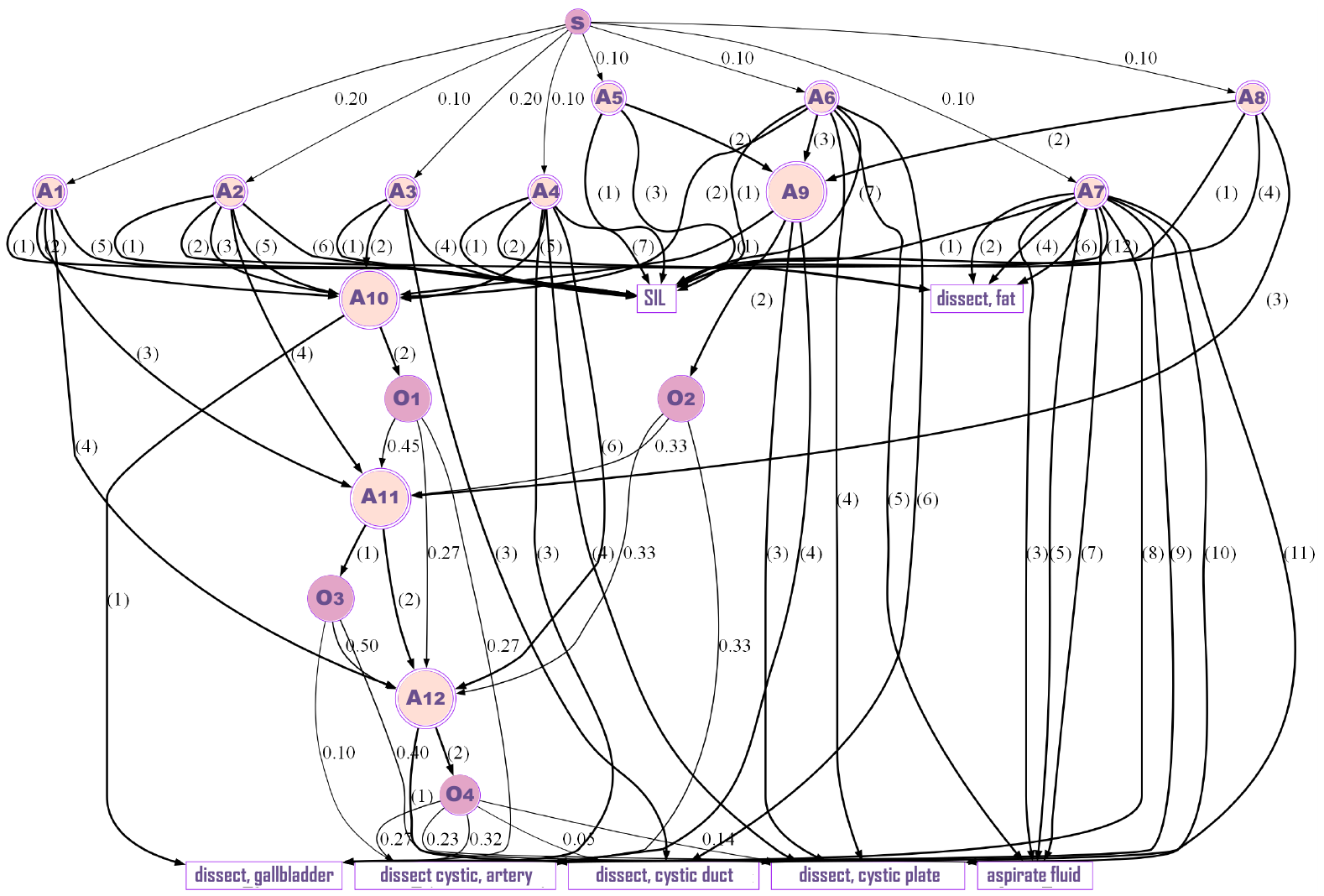} \vspace{-0.15in}
\caption{SP-AOG learned from a corpus of 10 surgeries. The pink nodes represent And-nodes, while the purple nodes signify Or-nodes. The numbers on the branching edges of Or-nodes indicate branching probability, and the bracketed numbers on And-node edges denote the order of expansion.} 
\label{Fig. learned PCFG} \vspace{-0.1in}
\end{figure}

We use the ADIOS algorithm for PCFG induction~\cite{qi2017predicting}. The divergence detection threshold in the representational data structure (RDS) graph is set to $\eta = 0.9$, with a significance test threshold of $\alpha = 0.08$. A context window size of 3 is applied, and the bootstrap threshold for equivalence classes is set at 0.65. The PCFGs learned from the corpus of 10 and 20 surgeries are denoted as $\mathcal{G}{10}$ and $\mathcal{G}{20}$, respectively. Fig. \ref{Fig. learned PCFG} presents a Graphviz visualization of $\mathcal{G}_{10}$.

\begin{figure*}[t] 
\centering 
\includegraphics[width=0.9\linewidth]{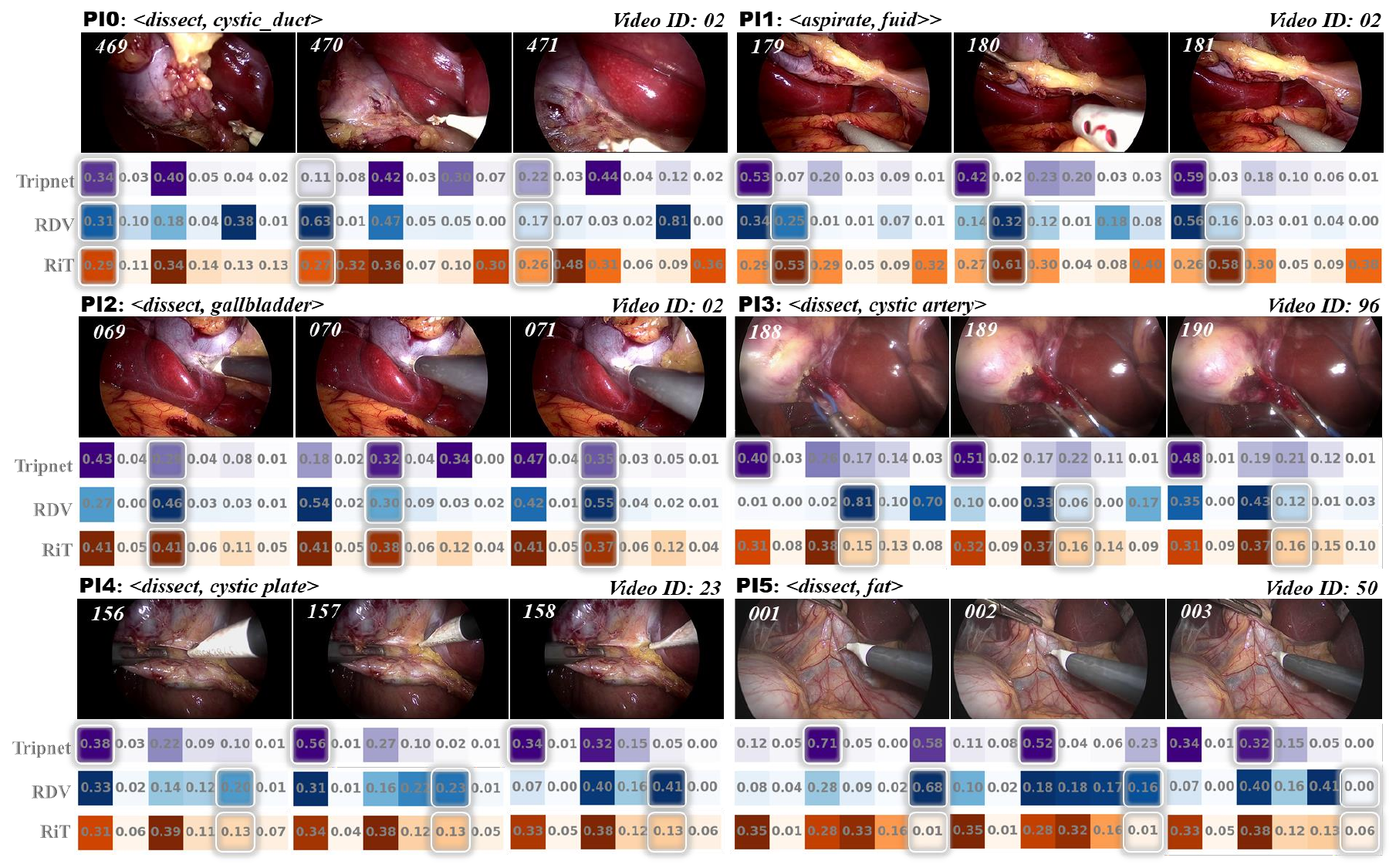} \vspace{-0.15in}
\caption{Visualization of refining PI recognition results from baseline models using surgical activity grammar. A grammar is induced to optimize Triplet, RDV, and RiT predictions for each PI class across sequential frames. The highest probability column in the probability matrix (columns represent PI0-PI6) indicates PI category predicted by each baseline model. White boxes highlight the PI categories that have been refined based on the surgical grammar.} 
\label{Fig. PI_case_improvement} \vspace{-0.15in}
\end{figure*}

\subsection{Experiment results}


\subsubsection{Quantitative results}
To assess the effectiveness of grammar-based refinement on surgical PI recognition, we replaced the classification model in the framework (Fig.~\ref{Fig. framework}) with a baseline model. In this way, the classification model is integrated with a grammar component to form the integration model, called "baseline+$\mathcal{G}$" (e.g., RDV+$\mathcal{G}$).
Furthermore, we explored the impact of PCFGs learned from training data of different sizes (denoted as $\mathcal{G}_{10}$ and $\mathcal{G}_{20}$) on PI recognition performance.
The quantitative results of our experiments are summarized in Table~\ref{tab:Performance Comparison}.
Our finding reveals that:

\begin{table} [t]
    \centering
   \caption{Performance Comparison: Baseline Methods vs. Grammar-Enhanced Approaches} \vspace{-0.1in}
    \label{tab:Performance Comparison}
    \begin{threeparttable}
     \begin{tabular}{cccccc}
    \toprule
       \multirow{2}{*}[-0.4ex]{\textbf{Methods}} & \textbf{{Micro}} & \multicolumn{3}{c}{\textbf{Macro}} &  \textbf{Weighted}\\
       \cline{3-5}
       \multirow{2}{*}{} & \textbf{P/R} & \textbf{Prec.} & \textbf{Recall} & \textbf{F1-score} & \textbf{F1-score}\\
       \hline
        Tripnet~\cite{nwoye2020recognition}  & 39.92 & 23.94 & 19.59 & 18.36 & 35.74\\ %
        Tripnet{\tiny W}~\cite{nwoye2020recognition}  & 35.86 & 19.47 & \textcolor{lightblue}{24.28} & 18.81 & 31.92\\  %
        \cellcolor{verylightgray} Tripnet+$\mathcal{G}_{10}$ & 43.04  & 20.56 & 19.74 & 17.76 & 37.59\\ %
        \cellcolor{verylightgray} Tripnet{\tiny W}+$\mathcal{G}_{10}$ & 43.16  & \textcolor{lightblue}{24.27} & 23.29 & \textcolor{lightblue}{21.25} & 35.58\\ %
        \cellcolor{verylightgray} Tripnet+$\mathcal{G}_{20}$ & \textcolor{lightblue}{43.23} & 19.40 & 20.05 & 18.15 & \textcolor{lightblue}{37.75}\\  %
        \cellcolor{verylightgray} Tripnet{\tiny W}+$\mathcal{G}_{20}$ & 42.21 & 22.86 & 22.02 & 20.10 & 35.03\\ %
        \hline
        RiT~\cite{sharma2023rendezvous}  & 35.62 & 23.05 & 19.92 & \textcolor{lightblue}{20.11} & 35.14\\ %
        RiT{\tiny W}~\cite{sharma2023rendezvous}  & 38.09 & 21.16 & 19.98 & 17.92 & 33.59\\ %
        \cellcolor{verylightgray} RiT+$\mathcal{G}_{10}$ & 35.81  & 30.58 & 18.97 & 18.77 & 34.92\\ %
        \cellcolor{verylightgray} RiT{\tiny W}+$\mathcal{G}_{10}$ & \textcolor{lightblue}{42.49}  & 32.19 & \textcolor{lightblue}{21.36} & 19.61 & \textcolor{lightblue}{36.46}\\ %
        \cellcolor{verylightgray} RiT+$\mathcal{G}_{20}$ & 36.67 & 24.27 & 19.71 & 19.86 & 35.65\\ %
        \cellcolor{verylightgray} RiT{\tiny W}+$\mathcal{G}_{20}$ & 41.88 & \textcolor{lightblue}{32.36} & 20.70 & 19.03 & 35.33\\ %
        \hline
        RDV~\cite{nwoye2022rendezvous}  & 41.47 & 32.46 & 28.98 & 29.91 & 39.75\\ %
        RDV{\tiny W}~\cite{nwoye2022rendezvous}  & 36.94 & 30.37 & 27.78 & 28.37 & 36.51\\ %
        \cellcolor{verylightgray} RDV+$\mathcal{G}_{10}$ & \textbf{\textcolor{blue}{44.52}} & 31.98 & 25.78 & 26.48 & \textbf{\textcolor{blue}{40.95}}\\ %
        \cellcolor{verylightgray} RDV{\tiny W}+$\mathcal{G}_{10}$ & 40.49 & 37.55 & 25.10 & 26.66 & 39.01\\ %
        \cellcolor{verylightgray} RDV+$\mathcal{G}_{20}$ & 43.72 & 41.27 & \textbf{\textcolor{blue}{30.10}} & \textbf{\textcolor{blue}{32.20}} & 40.56\\ %
        \cellcolor{verylightgray} RDV{\tiny W}+$\mathcal{G}_{20}$ & 39.80 & \textbf{\textcolor{blue}{46.66}} & 25.63 & 28.11 & 38.55\\ %
       \bottomrule
    \end{tabular}
    \begin{tablenotes}
        \footnotesize
        \item  \textcolor{lightblue}{Lightblue} indicates the best result related with each baseline model. 
        \item  \textcolor{blue}{Blue} highlights the overall best result across all methods.  
      \end{tablenotes}
    \end{threeparttable} \vspace{-0.25in}
\end{table}

\textbf{(1) Impact of Surgical Grammar}: Baseline models generally achieve better performance when incorporating learned surgical grammar. The RDV model surpasses the Tripnet and RiT models, and its performance improves even further when combined with grammar (RDV +$\mathcal{G}$).
Empirical evidence indicates that the PIs of the surgical procedure align well with the activity grammar of the SP-AOG. Leveraging this grammatical knowledge can significantly improve the performance of these baseline models that rely solely on visual features.

\textbf{(2) Impact of Weighted Loss}: Although models trained with weighted loss functions generally underperform compared to those trained with the original loss, they often produce predicted probability vectors with less bias towards high-proportion samples. This characteristic can be beneficial for grammatical parsing tasks, which could improve performance in certain scenarios, as demonstrated by the comparison between Tripnet+$\mathcal{G}_{10}$ and Tripnet{\tiny W}+$\mathcal{G}_{10}$.

Due to the significant class imbalance in the PI categories, models trained with weighted loss may be susceptible to overfitting the training set, resulting in suboptimal performance on the test set, as shown in the results of RDV model. 
However, incorporating grammars into these models can still improve performance on specific metrics, such as the Micro Macro Weighted P/R and the Weighted F1 score.

\textbf{(3) Impact of Grammar Size}: 
The relationship between the training data size used for grammar learning and its impact on PI recognition remains unclear. 
We did not assess the performance of grammars trained on larger datasets, as the temporal dynamics of surgical procedures for a 6-category PI set can be adequately captured from a dataset of 10-20 surgeries.

\subsubsection{Qualitative results}

To qualitatively evaluate the influence of grammar on PI classification, Fig.~\ref{Fig. PI_case_improvement} visually compares the predictions of baseline models with those incorporating surgical grammar. Fig.~\ref{Fig. PI_case_improvement} presents category probability matrices for each frame, where the highest probability column indicates the predicted PI category from baseline models. White boxes highlight the PI categories refined through our grammar-based approach. Key observations from the visualization result include:





\textbf{(1) Enhanced Accuracy}: Our grammar method consistently outperforms baseline models, producing more accurate predictions (white boxes) across all PI categories.

\textbf{(2) Error Correction}: Even when baseline models make intermittent error, our grammar-guided method can leverage temporal context to identify and correct these mistakes (e.g., PI0 in Fig.~\ref{Fig. PI_case_improvement}).

\textbf{(3) Temporal Importance}: When baseline models repeatedly mis-classify consecutive frames, these sequences are identified as ungrammatical, leading to the retention of initial predictions (e.g., PI1--Tripnet in Fig.~\ref{Fig. PI_case_improvement}). This highlights the critical role of temporal features in surgical action analysis.

In summary, by incorporating the hierarchical structure and temporal dependencies of surgical activities into a grammar-based framework, our method effectively mitigates the errors of baseline models in surgical behavior recognition.

\section{Conclusion} \label{Conclusion}

This paper presents a novel framework for recognizing primary intentions (PIs) in surgical procedures. By integrating DNN-based classification model with the surgical activity grammar, we effectively combine top-down procedural knowledge with bottom-up visual information. Our framework comprises three key modules: a top-down PI classifier, a grammar extractor, and a grammar parser. 
Our framework demonstrate a significant improvement over vision-based surgical activity detectors when evaluated on the benchmark dataset.


Future research will focus on harnessing the grammar-like structure of surgical activities to enhance automated surgical systems. Specifically, we aim to develop more robust and adaptable planning algorithms for autonomous surgical robots by leveraging the grammatical structure for high-level task planning.
Moreover, the parallels between surgical processes and natural language suggest that recent advancements in large language models (LLMs) for planning could be applied to surgical automation. We intend to investigate how LLMs, trained on surgical data, can generate human-understandable, robot-executable plans. The similarities between surgical procedures and natural language structure offer promising avenues for exploring procedure planning in the field of surgical automation.

\newpage
\bibliographystyle{unsrt}
\bibliography{ref}

\end{document}